\PassOptionsToPackage{table,dvipsnames}{xcolor}
\documentclass[sigconf, nonacm]{acmart}    
\usepackage{balance}      
\usepackage{stfloats}     
\usepackage{multirow}     
\usepackage{hyperref}     
\usepackage{comment}      
\usepackage{array}
\usepackage{makecell}     

\definecolor{lgray}{gray}{0.9}  
\definecolor{mgBaseline}{HTML}{4E79A7}   
\definecolor{mgNN}{HTML}{F28E2B}         
\definecolor{mgGNN}{HTML}{59A14F}        
\definecolor{mgFunctional}{HTML}{B07AA1} 
\definecolor{mgTree}{HTML}{E15759}       
\newif\ifmheadingcolor
\mheadingcolortrue  
\newcommand{\mcolor}[2]{\ifmheadingcolor{\setlength{\fboxsep}{1pt}\colorbox{#1!30}{#2}}\else#2\fi}

\usepackage{enumitem}       
\newcolumntype{H}{>{\setbox0=\hbox\bgroup}c<{\egroup}@{}}  
\usepackage[normalem]{ulem}  

\newcommand{\order}[1]{\mathcal{O}(#1)}     
\newcommand{\RR}{\mathbb{R}}                

\renewcommand{\vec}[1]{\boldsymbol{\mathrm{#1}}}  

\newcommand{\mat}[1]{\boldsymbol{#1}}         

\newcommand{\githuburl}{https://github.com/purduedb/shortest-distance-survey}  

\AtBeginDocument{%
 }

\begin{document}
\sloppy

\title{An Empirical Survey and Benchmark of Learned Distance Indexes for Road Networks}

\author{Gautam Choudhary}
\affiliation{%
 \institution{Purdue University}
 \city{West Lafayette}
 \state{IN}
 \country{USA}
}
\orcid{0000-0002-0261-7851}
\email{gchoudha@purdue.edu}

\author{Libin Zhou}
\affiliation{%
 \institution{Purdue University}
 \city{West Lafayette}
 \state{IN}
 \country{USA}
}
\orcid{0009-0006-3050-181X}
\email{zhou822@purdue.edu}

\author{Yeasir Rayhan}
\affiliation{%
 \institution{Purdue University}
 \city{West Lafayette}
 \state{IN}
 \country{USA}
}
\orcid{0000-0003-0326-2965}
\email{yrayhan@purdue.edu}

\author{Walid G. Aref}
\affiliation{%
 \institution{Purdue University}
 \city{West Lafayette}
 \state{IN}
 \country{USA}
}
\orcid{0000-0001-8169-7775}
\email{aref@purdue.edu}

\renewcommand{\shortauthors}{Choudhary et al.}

\begin{abstract}
The calculation of shortest-path distances in road networks is a core operation in navigation systems, location-based services, and spatial analytics. Although classical algorithms, e.g., Dijkstra's algorithm, provide exact answers, their latency is prohibitive for modern real-time, large-scale deployments. Over the past two decades, numerous distance indexes have been proposed to speed up query processing for shortest distance queries. More recently, with advances in machine learning (ML), researchers have designed and proposed ML-based distance indexes to answer approximate shortest path and distance queries efficiently. However, a comprehensive and systematic evaluation of these ML-based approaches is lacking. This paper presents the first empirical survey of ML-based distance indexes on road networks, evaluating them along four key dimensions: training time, query latency, storage, and accuracy.
Using 13 real-world road networks (7 with workload-driven queries, derived from trajectory data, and 6 with synthetically generated queries), we benchmark ten representative ML techniques and compare them against classical non-ML baselines, highlighting key insights and practical trade-offs. We release a unified open-source codebase to support reproducibility and future research on learned indexes.
\end{abstract}



\maketitle
{\renewcommand{\thefootnote}{}\footnotetext{Accepted at SIGSPATIAL '26. \url{https://doi.org/10.1145/3841645.3842992}}}

\section{Introduction}
Road networks are ubiquitous. The calculation of the shortest-path distance is a fundamental operation underlying a wide range of road-network applications, including navigation, route planning (e.g.,~\cite{route-planing-ref,route-planing-ref2,route-planing-ref3,route-planing-ref4}), traffic monitoring (e.g.,~\cite{traffic-monitoring-ref}), and location-based recommendation (e.g.,~\cite{spatial-crowd-sourcing-ref,poi-rec-ref}). While some of these applications involve time-dependent distances~\cite{wang2019querying,wagner2005geometric}, this survey only considers static edge weights.
Road networks are typically modeled as weighted graphs. Dijkstra's algorithm~\cite{dijkstra1959note} computes exact distances without preprocessing but is too slow for large-scale systems that operate over large graphs and serve millions of queries in real-time. Alternatively, precomputing all-pairs distances enables constant-time queries but is infeasible due to its $O(n^3)$ preprocessing time and $O(n^2)$ storage cost.

\noindent\textbf{Classical Distance Indexes.}
To navigate this space--time tradeoff, a large body of work has proposed preprocessing-based techniques that accelerate distance queries by storing auxiliary data in both general graphs and road networks.
Although the literature uses various terms (e.g., distance oracles, path oracles, and indexes), we use the term \emph{distance index} to refer to any such technique.
Over the past two decades, research has produced a wide range of \emph{classical distance indexes} that construct compact data structures to exploit structural properties of road networks, including graph hierarchies~\cite{HH-ref,ch-ref,CCH-Reference,UE-ref}, separators~\cite{silc2005-ref,pathOracle-ref,DO-2009-ref}, landmarks~\cite{sketch-do2-ref,landmarkA*-ref,billionNodeDO-ref}, and hub labeling~\cite{hub-based-labeling2010-ref,hierarchical-hub-labeling-ref,H2H-ref,HC2L-ref}. These methods achieve microsecond-level query latency~\cite{hub-based-labeling2010-ref,hierarchical-hub-labeling-ref,H2H-ref,HC2L-ref} by trading increased preprocessing time and storage for faster queries. Their performance may depend on the distance between the queried vertices~\cite{HH-ref,ch-ref,UE-ref,silc2005-ref}.

\noindent\textbf{Learned Distance Indexes.}
More recently, researchers have explored \emph{learned distance indexes} that use ML to approximate shortest-path distances rather than relying on explicit structural decompositions.
These approaches span neural networks~\cite{jindal2017unified,rizi2018shortest,qu2023learning,qi2020learning,chen2022ndist2vec}, graph neural networks~\cite{meng2024rgcndist2vec}, representation learning methods~\cite{kutuzov2019making,pacini2023aneda,zhao2022rne}, and tree-based models~\cite{jiang2021shortest}.
By accepting approximate answers, learned indexes trade exactness for reduced storage and simpler index structures, and leverage modern hardware accelerators, e.g., GPUs for training/inference.
In optimized settings, as shown in this paper, these models can achieve low query latency---down to nanoseconds, significantly faster than classical distance indexes (at the cost of approximation error and training overhead).

\noindent\textbf{Research Gap.}
Despite growing interest, the space of learned distance indexes has received little attention, and specifically, in the context of road networks. No prior work has systematically evaluated ML-based distance indexes in the road-network setting.
Existing surveys focus on shortest-path algorithms~\cite{shortest-path-survey-2014-ref,shortest-path-survey-2017-ref} or classical distance indexes~\cite{do-analytics-ref,2013-query-evaluation-ref,2017-hub-labeling-evaluation,dynamic-sp-evaluation-ref}.
Recent evaluations on general unweighted graphs~\cite{do-evaluation-2022-ref} provide a thorough comparison of node embedding approaches for distance estimation, but their scope does not extend to the learned distance indexes considered in this paper.
This gap is important because road networks differ fundamentally from social or information networks. They are sparse, nearly planar, and exhibit low and near-uniform degree distributions but much larger diameters~\cite{road-properties,snap-ref}. These structural properties can be leveraged to accelerate distance computation, e.g., California road network has 1.9M vertices with an Average Degree of 1.4 and Diameter 849, whereas the Pokec social network of comparable size has an Average Degree of 18.8 and Diameter 11.

\noindent\textbf{Objectives.} Learned distance indexes introduce their own challenges and tradeoffs. A learned index must balance accuracy, query latency, storage footprint, training cost, and scalability to large road networks. This paper systematically evaluates existing learned indexes along these dimensions and compares them against classical baselines to analyze their tradeoffs, relative strengths, and weaknesses.

The contributions of this paper are as follows:
\begin{itemize}[leftmargin=*]
\item \textbf{Empirical Survey of Learned Distance Indexes.}
We present the first systematic survey and experimental study of learned distance indexes for shortest-path distance estimation on road networks. Our study covers 10 ML methods spanning NNs, GNNs, and gradient-boosted trees, and contrasts them with three non-ML baselines, evaluated across 13 real-world road networks.

\item \textbf{Encoder--Decoder Abstraction.}
We introduce a unified encoder--decoder abstraction that structures the design space of learned distance indexes and provides a common conceptual framework for understanding existing and future methods.

\item \textbf{Workload-driven Benchmark and Key Insights.}
We introduce a workload-driven benchmark derived from real queries alongside synthetic queries used in prior studies, to reflect practical use in real-world settings. The benchmark evaluates key dimensions---approximation error, preprocessing time, query latency, and storage overhead---and reveals tradeoffs that guide the design and comparison of learned and non-ML distance indexes.

\item \textbf{Open-source.}
We release a unified open-source evaluation framework for benchmarking learned distance indexes, addressing the sparsity of existing implementations and enabling reproducibility and integration of new techniques. {Code is available at \url{\githuburl}}
\end{itemize}

The paper proceeds as follows. Section~\ref{sec:preliminaries} formalizes the problem and introduces background material. Section~\ref{sec:methods} surveys ML-based distance indexes. Section~\ref{sec:experiments} presents experimental evaluation. Section~\ref{sec:related-work} reviews related work, and Section~\ref{sec:conclusion} concludes the paper.

\section{Preliminaries}
\label{sec:preliminaries}
\noindent\textbf{Problem Definition.}
We model a road network as an undirected weighted graph $G = (\mathcal{V}, \mathcal{E}, w)$, where $\mathcal{V}$ is the set of $n$ vertices (or nodes) representing road intersections, and $\mathcal{E}$ is the set of $m$ edges representing road segments.
Each edge $(u,v) \in \mathcal{E}$ has a positive weight $w(u,v) \in \RR_{+}$ denoting road distance between $u$ and $v$. Each vertex $v \in \mathcal{V}$ is also associated with geographic coordinates (latitude and longitude).
A path $P(s,t)$ from Vertex $s$ to $t$ is an ordered sequence $\langle v_0, v_1, \dots, v_x \rangle$ such that $(v_i, v_{i+1}) \in \mathcal{E}$, $v_0 = s$, and $v_x = t$. The length (or cost) of a path is
$
|P(s,t)| = \sum_{i=0}^{x-1} w(v_i,v_{i+1}).
$
The shortest path $SP(s,t)$ is the path minimizing this total cost, and the corresponding \emph{shortest distance query} returns only its length $d(s,t) = |SP(s,t)|$.
\textit{Approximate} shortest distance returns an estimate $\hat{d}(s,t)$ that may deviate slightly from the true distance $d(s,t)$.

An \emph{ML distance index} trains a model on vertex-pair samples $\mathcal{D}=\{(u_i, v_i, d_{u_i v_i})\}_{i=1}^{N}$ and learns low-dimensional embeddings for vertices. The model approximates the true distance function, $\hat{d}(u,v) \approx d(u,v)$, thereby compressing the full $O(n^2)$ distance matrix into $O(nd)$ space, where $d$ is the embedding dimension. At query time, the trained model estimates $\hat{d}(u,v)$ directly from the vertex embeddings and model parameters. Let $d_{uv}$ and $\hat{d}_{uv}$ be shorthands for $d(u,v)$ and $\hat{d}(u,v)$ respectively.
This paper focuses on \emph{distance queries}---rather than path reconstruction---and evaluates ML-based distance indexes across several dimensions.

\noindent\textbf{Neural Networks (NN).}
An NN is a parameterized, differentiable function composed of multiple layers of linear transformations followed by non-linear activations. NNs are \emph{universal function approximators}. Given sufficient capacity and appropriate loss functions, they can approximate a function to any desired accuracy. For a single hidden layer, the transformation is written as:
\begin{equation}
\label{eq:nn}
\mat{H}_{i} = f_{\text{NN}}(\mat{H}_{i-1}; \mat{W}_i, \mat{X}) = \sigma(\mat{H}_{i-1} \mat{W}_i),
\end{equation}
where $\mat{H}_{i-1}$ and $\mat{H}_{i}$ are the input and output activations, $\mat{W}_i$ are the learnable parameters, and $\sigma(\cdot)$ is a non-linear activation function, e.g., ReLU or tanh. The input layer is $\mat{H}_0 = \mat{X}$, where $\mat{X}$ is the input feature matrix, either learned during training or initialized from pretrained embeddings. The NN is trained as a regression model to approximate shortest distances between node pairs, i.e., $f_{\text{NN}}(u,v) \approx d(u,v)$ where the embeddings of $u$ and $v$ are fused (e.g., concatenated) to form a single input vector to the network.

\noindent\textbf{Graph Neural Networks (GNN).}
GNNs extend traditional NNs to graph-structured data by incorporating the topological dependencies and connectivities among vertices. A GNN layer updates node representations through message passing expressed as:
\begin{equation}
\label{eq:gnn}
\mat{H}_{i} = f_{\text{GNN}}(\mat{H}_{i-1}; \mat{A}, \mat{W}_i, \mat{X}) = \sigma(\mat{A} \mat{H}_{i-1} \mat{W}_i),
\end{equation}
where $\mat{A}$ is the (normalized) adjacency matrix encoding graph connectivity, $\mat{H}_i$ represents node features at Layer $i$, $\mat{W}_i$ is the layer-specific weight matrix, and $\sigma(\cdot)$ is a non-linear activation. Initially, $\mat{H}_0 = \mat{X}$, either learned or pretrained. By repeated propagation, GNNs aggregate information from multi-hop neighbors, enabling them to learn representations that capture both spatial proximity and road network structural patterns to estimate pairwise distances.

\begin{table*}[t]
\caption{Encoder and decoder architectures for all models evaluated. \colorbox{lgray}{Shaded} labels denote learnable modules. \\
RN=Road Network, SN=Social Network, KG=Knowledge Graph, QT=Query Time, MS=Model Size, SU=Supports Updates.}
\label{tab:encoder-decoder-summary}
\small
\begin{tabular}{
    l|  
    c|  
    c|  
    >{\centering\arraybackslash}p{1.0cm}|  
    >{\centering\arraybackslash}p{1.2cm}|  
    >{\centering\arraybackslash}p{1.0cm}|  
    >{\centering\arraybackslash}p{1.6cm}   
}
\toprule
\textbf{Model}
 & \textbf{Encoder}
 & \textbf{Decoder}
 & \textbf{QT}
 & \textbf{MS}
 & \textbf{SU}
 & \textbf{Domain} \\
\midrule
\multicolumn{7}{c}{\textbf{Baselines}} \\
Manhattan
 & Coordinates ($d=2$)
 & $\ell_1$-norm
 & $\order{1}$
 & $\order{n}$
 & $\times$
 & -- \\
Landmark~\cite{takes2014adaptive, potamias2009fast}
 & Landmark distance vectors
 & minimum
 & $\order{d}$
 & $\order{nd}$
 & $\times$
 & RN and SN \\
\midrule
\multicolumn{7}{c}{\textbf{Neural Network (NN) Methods}} \\
GeoDNN~\cite{jindal2017unified}
 & Coordinates ($d=2$)
 & \cellcolor{lgray}{NN($2d$, 20, 100, 20, 1)}
 & $\order{1}$
 & $\order{n}$
 & $\times$
 & RN \\
DistanceNN~\cite{rizi2018shortest}
 & Pretrained embeddings
 & \cellcolor{lgray}{NN($d$, 64, 12, 1)}
 & $\order{d}$
 & $\order{nd}$
 & $\times$
 & SN \\
EmbedNN~\cite{qu2023learning}
 & Pretrained embeddings
 & \cellcolor{lgray}{NN($d$, 500, 1)}
 & $\order{d}$
 & $\order{nd}$
 & $\checkmark$
 & SN \\
Vdist2vec~\cite{qi2020learning}
 & \cellcolor{lgray}{Learnable embeddings}
 & \cellcolor{lgray}{NN($2d$, 100, 20, 1)}
 & $\order{d}$
 & $\order{nd}$
 & $\times$
 & RN \\
Ndist2vec~\cite{chen2022ndist2vec}
 & \cellcolor{lgray}{Learnable embeddings}
 & \cellcolor{lgray}{$4{\times}$NN($2d$, 100, 20, 1)}
 & $\order{d}$
 & $\order{nd}$
 & $\times$
 & RN \\
LandmarkNN
 & Landmarks + coordinates
 & \cellcolor{lgray}{NN($2d$, 1024, 512, 1)}
 & $\order{d}$
 & $\order{nd}$
 & $\times$
 & -- \\
\midrule
\multicolumn{7}{c}{\textbf{Graph Neural Network (GNN) Methods}} \\
RGCNdist2vec~\cite{meng2024rgcndist2vec}
 & \cellcolor{lgray}{GCN(2, 512, $d$)}
 & $\ell_1$-norm
 & $\order{d}$
 & $\order{n+d}$
 & $\times$
 & RN \\
RSAGEdist2vec
 & \cellcolor{lgray}{SAGE(2, 512, $d$)}
 & $\ell_1$-norm
 & $\order{d}$
 & $\order{n+d}$
 & $\times$
 & -- \\
RGATdist2vec
 & \cellcolor{lgray}{GAT(2, 512, $d$)}
 & $\ell_1$-norm
 & $\order{d}$
 & $\order{n+d}$
 & $\times$
 & -- \\
\midrule
\multicolumn{7}{c}{\textbf{Functional Methods}} \\
Path2Vec~\cite{kutuzov2019making}
 & \cellcolor{lgray}{Learnable embeddings}
 & cosine distance
 & $\order{d}$
 & $\order{nd}$
 & $\times$
 & KG \\
ANEDA~\cite{pacini2023aneda}
 & \cellcolor{lgray}{Learnable embeddings}
 & cosine distance
 & $\order{d}$
 & $\order{nd}$
 & $\times$
 & RN \\
RNE~\cite{zhao2022rne}
 & \cellcolor{lgray}{Hierarchical embeddings}
 & $\ell_1$-norm
 & $\order{d}$
 & $\order{nd}$
 & $\times$
 & RN \\
\midrule
\multicolumn{7}{c}{\textbf{Gradient-Boosted Decision Tree (GBDT) Methods}} \\
CatBoost~\cite{jiang2021shortest}
 & Landmarks + coordinates
 & \cellcolor{lgray}{GBDT}
 & $\order{d+c}$
 & $\order{nd+c}$
 & $\times$
 & RN \\
\bottomrule
\end{tabular}
\end{table*}

\section{Methods}
\label{sec:methods}

\begin{figure}[h]
\centering
\includegraphics[width=0.99\linewidth]{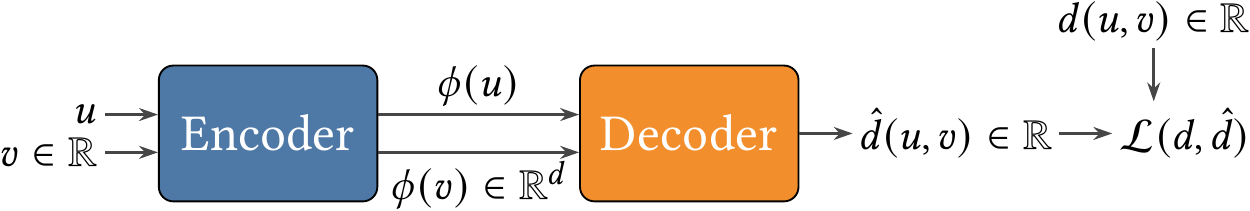}
\caption{Encoder-decoder for shortest-distance estimation.
}
\label{fig:encoder-decoder}
\end{figure}

Learned distance indexes can be viewed in a unified \emph{encoder-decoder} framework. Given a node pair, the encoder maps each node to a vector, and the decoder maps the pair of vectors to a distance estimate. Figure~\ref{fig:encoder-decoder} illustrates this abstraction.
The \textbf{encoder} is a function $\phi:\mathcal{V}\rightarrow\RR^d$ that maps a node $v\in\mathcal{V}$ to a $d$-dimensional embedding vector $\vec{h}_v=\phi(v)$. This mapping can be realized simply by lookup tables or by GNNs that capture structural and spatial dependencies within the graph. Lookup tables may store either fixed representations, e.g., coordinates or landmark distances, or trainable embeddings. The resulting embeddings for all nodes form an $n\times d$ matrix.
The \textbf{decoder} is a function $\psi:\RR^d \times \RR^d \rightarrow \RR_{\ge 0}$ that takes a pair of node embeddings $\vec{h}_u$ and $\vec{h}_v$ and produces a scalar estimate $\hat{d}_{uv}=\psi(\vec{h}_u,\vec{h}_v)$ of the shortest-path distance between the corresponding nodes. Decoding involves a \emph{fusion operator} to combine the embeddings into a joint representation (e.g., concatenation), and a \emph{scoring function} to map the representation to a distance value. The scoring function can be a fixed form, e.g., an $\ell_1$ or $\ell_2$ norm, or a parameterized model, e.g., an NN. Fusion operators include concatenation ($\vec{h}_u \oplus \vec{h}_v$), average ($(\vec{h}_u + \vec{h}_v)/2$), sum ($\vec{h}_u + \vec{h}_v$), difference ($\vec{h}_u - \vec{h}_v$), or Hadamard product ($\vec{h}_u \odot \vec{h}_v$).

Next, we present existing models using this encoder--decoder abstraction. Each method is discussed along three dimensions: scalability to large graphs, adaptability to evolving networks, and computational complexity. Shared properties are discussed below.

\noindent\textbf{Time, Space, and Updates.}
Generally, all methods store a $d$-dimensional embedding per node in the encoder and learnable weights in the decoder. Most parameters reside in the encoder's embedding matrix, scaling linearly with embedding dimension $d$ and number of nodes $n$. Decoder parameters contribute negligibly: at most $\mathcal{O}(d)$ for a neural decoder or zero for a functional one. Hence, model size scales as $\mathcal{O}(dn)$, or $\mathcal{O}(n)$ for a fixed $d$.
Query latency depends on the decoder complexity, since the encoder is a fast embedding lookup. Inference involves no graph traversal. This gives $\mathcal{O}(d)$ query time, effectively $\mathcal{O}(1)$ for a fixed architecture. Functional decoders (e.g., $\ell_p$-norm) are highly efficient and faster than NN-based decoders, making them preferable for scalable distance estimation.
Most methods do not support incremental updates (e.g., add or delete edges) and hence require retraining upon updates.

\vspace{3pt}
\noindent\textbf{\mcolor{mgBaseline}{M0. Landmarks (Non-ML)}~\cite{takes2014adaptive, potamias2009fast}.}
Although non-ML in nature, the landmark-based approach fits naturally into the encoder--decoder abstraction and serves as a useful reference point for subsequent learned indexes.
The key idea is to precompute distances from every node to a small subset of nodes, termed \emph{landmarks}, and use these as a compact surrogate for the full distance matrix. Selecting the optimal landmarks is \textbf{NP-hard}~\cite{potamias2009fast}. Heuristics using node centrality, e.g., degree, closeness, and betweenness are used.
The encoder corresponds to a fixed $n \times l$ embedding matrix $\mat{H}$, where each entry $H_{ij}$ stores the shortest distance between Node $i$ and Landmark $j$, and $l$ is the number of selected landmarks (equal to the embedding dimension $d$). The decoder estimates the distance between two nodes by minimizing over landmark indices: $\hat{d}_{uv} = \psi(\vec{h}_u, \vec{h}_v) = \min_i \big(\phi(u)_i + \phi(v)_i\big)$. This formulation follows from the triangle inequality that bounds the true distance $d_{uv}$ as
$
|d_{ui} - d_{iv}| \le d_{uv} \le d_{ui} + d_{iv}, \quad \forall i \in \mathcal{V}.
$
Several variants use alternative estimators within this range (e.g., lower bound, mean, or geometric mean), but the upper bound is most commonly adopted due to its empirical accuracy.
As no learning is involved, this method only requires index construction to compute shortest distances from each landmark to all nodes, typically via BFS or Dijkstra traversals.
Landmark distances can be computed in parallel, scaling well to large graphs. Unlike learned methods, network changes require recomputing landmark distances rather than retraining.

\vspace{3pt}
\noindent\textbf{\mcolor{mgNN}{M1. GeoDNN}~\cite{jindal2017unified}.}
As the earliest NN-based approach for distance estimation, GeoDNN adopts a unified NN architecture, originally proposed as ST-NN, for jointly estimating travel time and distance in taxi-trip data.
We focus on its distance sub-module \emph{DistNN} that predicts the travel distance between a source and destination. The fixed geo-coordinates serve as node features for the encoder, defined as $\phi(u) = \vec{h}_u = [\text{long}_u, \text{lat}_u]$. For a given node pair $(u,v)$, the corresponding embeddings are concatenated using the \emph{Concat} fusion operator and
are
passed to a four-layer feed-forward NN that acts as the decoder. As the model operates directly on geographic coordinates, it is referred to as GeoDNN.
Formally,
$
\hat{d}(u,v) = f_{\text{NN}}(\vec{h}_u \oplus \vec{h}_v),
$
where $\vec{h}_u$ and $\vec{h}_v$ represent the fixed coordinate embeddings of Nodes $u$ and $v$, respectively. The model is trained using the mean squared error loss (MSE). GeoDNN is evaluated on the NYC Taxi dataset that contains pickup and dropoff coordinates, timestamps, trip distances, and durations.
With $d{=}2$ (lat/long coordinates), space reduces to $\mathcal{O}(n)$ and query time to $\mathcal{O}(1)$. No mechanism is proposed for scaling to large graphs.

\vspace{3pt}
\noindent\textbf{\mcolor{mgNN}{M2. DistanceNN}~\cite{rizi2018shortest}}
is among the first to leverage pre-trained graph embeddings for supervised distance prediction. Node embeddings are first learned in an unsupervised manner using graph embedding, e.g., Node2Vec~\cite{grover2016node2vec} or Poincar\'e embeddings~\cite{nickel2017poincare}, and are frozen to serve as the encoder $\phi(u)=\vec{h}_u$. The decoder is a three-layer feed-forward NN trained to predict the shortest-path distance for node pairs after combining the two embeddings with a fusion operator (e.g., average $(\vec{h}_u + \vec{h}_v)/2$). DistanceNN is similar to GeoDNN in structure, but differs in its input features--graph embeddings vs. geo-coordinates.
Node2Vec embeddings along with the \emph{Average} fusion operator are found to yield accurate results. Testing is conducted on social network graphs, where the maximum shortest-path length spans only few hops.
To handle large graphs, landmark-based strategies confine training dataset size to $\mathcal{O}(nl)$.

\vspace{3pt}
\noindent\textbf{\mcolor{mgNN}{M3. EmbedNN}~\cite{qu2023learning}.}
As in DistanceNN, node embeddings are learned in an unsupervised manner from graph structure using Node\-2Vec \cite{grover2016node2vec} or LINE~\cite{tang2015line}, and is frozen to serve as the encoder $\phi(u)=\vec{h}_u$.
The decoder is a lightweight 2-layer feed-forward NN that predicts shortest-path distances among nodes after fusing their embeddings using a fusion operator, e.g., Hadamard product $\vec{h}_u \odot \vec{h}_v$.
DistanceNN and EmbedNN are evaluated on graphs, e.g., social networks, where the maximum shortest-path length rarely exceeds five to six hops.
EmbedNN investigates regression and classification formulations of distance prediction and reports that Node2Vec with Hadamard fusion yield better performance.

\noindent\textbf{Handling Large Networks and Updates.}
Although no strategy is proposed for scaling to large networks, the model can in principle handle them as long as the underlying embedding algorithm can be executed. EmbedNN has a heuristic for handling incremental updates. For each modified edge $(s,t) \in \mathcal{E}'$, the predicted distance $\hat{d}_{uv}$ for a query pair $(u,v)$ is compared to an alternative route passing through $(s,t)$, computed as $\hat{d}^*_{uv}=\hat{d}_{us}+d_{st}+\hat{d}_{tv}$, and the smaller of the two is returned. This heuristic filters out irrelevant updates, where modified edges are spatially distant from the queried path. But, it incurs overhead proportional to the number of modified edges ($2|\mathcal{E}'|$ extra predictions for $\hat{d}_{us}$ and $\hat{d}_{tv}$). The same idea can be readily extended to other distance indexes.

\vspace{3pt}
\noindent\textbf{\mcolor{mgNN}{M4. Vdist2vec}~\cite{qi2020learning}.}
Vdist2vec is the first end-to-end trainable model for shortest-path distance prediction, where the encoder and decoder are unified and optimized jointly. Unlike previous methods that use fixed or pre-trained embeddings, Vdist2vec learns node embeddings while training. Each node $u$ is associated with a learnable embedding vector $\vec{h}_u = \phi(u) \in \RR^d$ that is concatenated with that of another node to form the pair representation $\vec{h}_u \oplus \vec{h}_v$.
This fused representation is passed to a three-layer feed-forward NN decoder that outputs the predicted distance:
$\hat{d}(u,v) = f_{\text{NN}}(\vec{h}_u \oplus \vec{h}_v).$
The model is trained end-to-end using MSE loss with all encoder parameters (node embeddings) and decoder weights optimized jointly.
To improve prediction robustness, two variants are proposed.
\emph{Vdist2vec-L} replaces MSE loss with Huber loss to make it robust to outliers, as the threshold parameter $\delta$ is adaptively set to the top 1\% largest prediction errors after each epoch.
\emph{Vdist2vec-S} adopts an ensemble-style architecture by replacing the decoder network with 4 parallel sub-networks that share the same input features, each targeted for a specific distance range (short, mid, or long). The outputs of these sub-networks are aggregated to produce final distance estimates. This design improves prediction accuracy across distance scales.

\noindent\textbf{Handling Large Networks.}
To scale training to larger graphs, a clustering-based sampling strategy is proposed that reduces the quadratic all-pairs dataset to $\mathcal{O}(nl)$ training samples. Nodes are partitioned into $l \ll n$ clusters based on geographic proximity, and the model is trained to predict distances between each node and its cluster center.
At query time, the predicted distance between arbitrary nodes $(u,v)$ is approximated as:
$\hat{d}(u,v) = \lambda_1 \hat{d}(u, c_u) + d(c_u, c_v) + \lambda_2 \hat{d}(v, c_v),
$
where $c_u$ and $c_v$ are the nearest cluster centers of $u$ and $v$, respectively, $d(c_u, c_v)$ is the precomputed inter-cluster distance between these centers, and $\lambda_1$, $\lambda_2$ are learnable weighting factors. This strategy reduces training cost while maintaining high prediction accuracy. Though presented within Vdist2vec, this clustering-based scaling generalizes to other learned distance indexes.

\vspace{3pt}
\noindent\textbf{\mcolor{mgNN}{M5. Ndist2vec}~\cite{chen2022ndist2vec}.}
Ndist2vec extends Vdist2vec-S to improve training efficiency and scalability for large road networks. The model retains the same encoder-decoder structure as Vdist2vec-S, comprising learnable node embeddings and a multi-branch decoder, but differs in two key aspects; learning aggregation parameters and training strategy. Unlike Vdist2vec-S,
Ndist2vec learns branch aggregation weights jointly with the model, allowing data-driven calibration across distance ranges.
For training, Ndist2vec uses hybrid data generation that combines all-pairs and adaptive landmark-based sampling. The model is initially trained on all node pairs ($\mathcal{O}(n^2)$ samples) to capture global structure. Then, the training data is restricted to node-landmark pairs ($\mathcal{O}(nl)$ samples), where the landmark set is randomly refreshed in each epoch to ensure broader coverage while reducing training cost. This lowers the training burden with marginal loss in accuracy, making it practical for large-scale road networks.
Ndist2vec reduces training time by over 75\% while incurring minor degradation in accuracy. Across datasets, MRE increases by roughly 1--2\% compared to Vdist2vec.

\vspace{3pt}
\noindent\textbf{\mcolor{mgFunctional}{M6. Path2vec}~\cite{kutuzov2019making}.}
Path2vec learns dense node embeddings that preserve user-defined graph similarity or distance measures $sim: \mathcal{V} \times \mathcal{V} \rightarrow \RR$.
It generalizes naturally to graph metrics, e.g., shortest-path distance. Path2vec differs from previous neural approaches by using a (fixed) functional decoder instead of a neural one, enabling substantially faster inference.
The encoder consists of a learnable embedding matrix $\vec{H} \in \RR^{n \times d}$, where each node $u$ is assigned an embedding $\vec{h}_u = \phi(u)$. Embeddings are optimized using a modified loss function that combines distance reconstruction with a neighborhood-similarity regularizer:
\begin{equation}
\label{eq:path2vec-loss}
\mathcal{L} = \sum_{(u, v) \in \mathcal{D}} \Big[ (\hat{d}_{uv} - d_{uv})^2 - \alpha(\vec{h}_u^\top \vec{h}_i + \vec{h}_v^\top \vec{h}_j) \Big],
\end{equation}
where $i$ and $j$ are randomly sampled neighbors of $u$ and $v$, respectively, and $\alpha$ controls the regularization strength. The first term minimizes reconstruction error, and the second encourages neighboring nodes to have similar embeddings, nudging the optimization appropriately. Originally, the decoder computes similarity $\hat{d}_{uv}$ as a dot product,
$\vec{h}_u^\top \vec{h}_v$.
For distance estimation, we modify it to use cosine distance:
$\hat{d}_{uv} = \big(1 - \hat{\vec{h}}_u^\top \hat{\vec{h}}_v\big) \frac{d_{\max}}{2}$,
where
$\hat{\vec{h}}_u = \vec{h}_u / \|\vec{h}_u\|$ and $d_{\max}$
is the maximum observed graph distance.
This design resembles skip-gram models~\cite{mikolov2013efficient} (capturing local structure) and GloVe~\cite{pennington2014glove} embeddings (capturing global structure), offering a balance between locality and globality in graph representation learning.
For scalable training, Path2vec retains only top-$k$ (50--100) most similar node pairs per node, keeping the dataset size linear in $n$.

\vspace{3pt}
\noindent\textbf{\mcolor{mgFunctional}{M7. ANEDA}~\cite{pacini2023aneda}.}
ANEDA extends Path2vec with a simpler learning objective and systematic evaluation of functional distance measures. The encoder has a learnable embedding matrix $\mathbf{H} \in \RR^{n \times d}$, where each node $u$ is assigned an embedding $\vec{h}_u = \phi(u)$. The decoder is non-parametric, using analytical distance functions, e.g., $\ell_p$-norms, hyperbolic, elliptic, Minkowski, and inverse dot-product (i.e., cosine distance) measures to compute pairwise distances between embeddings.
Unlike Path2vec, ANEDA trains embeddings using MSE loss and explores various initializations, e.g., Node2Vec~\cite{grover2016node2vec}, GraRep~\cite{cao2015grarep}, and geo-coordinates.
Node2Vec initialization shows the most consistent gains. Random initialization is also good. The simplest inverse dot-product decoder is the best on road network datasets, followed by the $\ell_6$-norm. ANEDA has results comparable to Vdist2vec-S, but its accuracy is slightly lower due to the absence of a learnable decoder.
All-pairs query generation restricts experiments to small graphs. Large-scale training is not addressed.

\vspace{3pt}
\noindent\textbf{\mcolor{mgFunctional}{M8. RNE}~\cite{zhao2022rne}.}
Road Network Embedding (RNE) uses hierarchical embedding to partition the road network into progressively smaller subgraphs, forming a multi-level hierarchy (e.g., city, regional, and local). Each level is trained sequentially using simple MSE loss starting from the top level and down until each node is its own partition. The encoder uses the embeddings learned at each level. The decoder computes pairwise distances using $\ell_1$-norm. Hierarchical training captures the road network global and local structural features, with partitions encoding connectivity across spatial scales - a critical aspect for accurate distance estimation. After training, embeddings from all hierarchical levels are aggregated (e.g., summed across parent partitions) to produce a single embedding per node for inference. RNE combines elegant training with simple querying, hence is very effective among learned distance estimators.
RNE uses three-stage sampling to generate good node pairs, followed by active fine-tuning for efficient retraining in large networks.

\vspace{3pt}
\noindent\textbf{\mcolor{mgTree}{M9. CatBoost}~\cite{jiang2021shortest}.}
CatBoost uses gradient-boosted decision trees instead of fixed neural architectures.
Unlike previous models whose structure remains static once defined, CatBoost constructs its model by sequentially adding trees during training. New trees are trained to reduce residual errors of the previous ensemble. The encoder consists of fixed, hand-crafted features from the node pairs: landmark distances, geo-coordinates, Euclidean distance between coordinates, and cosine similarity of landmark vectors.
Landmark distances and coordinates are global features, while similarity-based features capture local pairwise relationships. The decoder is a two-stage CatBoost regressor. The first stage output serves as an additional feature for the second stage's final prediction.
CatBoost uses a landmark query generation strategy to handle large networks.

\noindent\textbf{Time and Space Complexity.}
Given node coordinates and $l$ landmark distances as input features, the model size is $\order{(l + 2)n}$, where the additional 2 is for latitude and longitude features. Other pairwise features (e.g., Euclidean distance, cosine similarity) are computed on the fly during inference. Query latency has two components: feature computation $\mathcal{O}(l + 2)$ and CatBoost inference cost $\mathcal{O}(c)$, where $c$ is the number of trees evaluated per query.

\vspace{3pt}
\noindent\textbf{\mcolor{mgGNN}{M10. RGCNdist2vec}~\cite{meng2024rgcndist2vec}.}
RGCNdist2vec uses a GNN for shortest distance estimation. It replaces the embedding lookup with a relational graph convolutional network (GCN)~\cite{kipf2016semi} that learns node embeddings directly from graph structure. The encoder follows the GNN formulation in Eqn~\eqref{eq:gnn}, propagating and aggregating information across neighboring nodes through multiple layers. After $L$ layers of message passing, each node obtains an embedding $\phi(u)=\mat{H}_L(u)\in\RR^d$. Embeddings of a node pair $(u,v)$ are passed to a lightweight decoder to estimate distance using $\ell_1$-norm: $\hat{d}_{uv}=||\phi(u)-\phi(v)||_1$. The model is trained using Smooth L1 loss that combines robustness of $\ell_1$ loss to outliers with the stability of $\ell_2$ loss for small residuals (errors).
Three-stage sampling is proposed to partition the graph to reduce training data and improve scalability.

\noindent\textbf{Time and Space Complexity.}
The GNN weights and input features (e.g., node coordinates and adjacency list) are stored instead of the full embedding matrix, reducing storage from $\order{nd}$ to $\order{n+d}$.
During inference, node embeddings are reconstructed on demand using the stored features and cached, amortizing their computation cost. This lightweight decoder design has query latency of $\mathcal{O}(d)$.

\subsection{Query Dataset Preparation}
Having introduced the model architectures, next, we describe how the supervision data, i.e., node pairs $(u,v)$ and their corresponding ground-truth shortest-path distances $d_{uv}$, are prepared from road networks for training and evaluating the models. The ground-truth distances $d_{uv}$ can be obtained using a state-of-the-art exact distance index.
The main design choice lies in how to select the node pairs $(u,v)$ that form the training and test query set. Queries are either \emph{workload-driven} derived from real-world trip data, e.g., taxi trips and GPS logs~\cite{nyc-taxi-ref,chicago-taxi-ref,geolife-ref}, or \emph{synthetically} generated by sampling strategies.

\subsubsection{\textbf{Workload-driven Sampling}}
Each trajectory contributes one or more queries by mapping pickup and drop-off locations to their nearest road-network nodes. This reflects the inherent bias of real systems, where certain routes are queried more frequently.
Only a portion of the full distance matrix is used, focusing learning on frequently traveled routes.
GeoDNN~\cite{jindal2017unified} uses a similar idea. Most prior studies rely on synthetic workloads. Our study is among the first workload-driven comparisons of learned distance indexes.

\subsubsection{\textbf{Synthetic Sampling}}
\label{sec:query-dataset-prep}
When real query workloads are unavailable, node pairs can be synthetically generated to approximate the underlying query distribution. Common strategies include:

\noindent\textbf{All-pairs:} It exhaustively computes distances for every node pair, producing $|\mathcal{V}|^2$ samples that form the full distance matrix.
This strategy is intuitive and straightforward for small networks. It assumes all nodes are equally important and provides complete coverage. Its quadratic growth makes it impractical for large networks.

\noindent\textbf{Landmark-based:} It selects a small set of landmark nodes and computes distances between all nodes and each landmark.
This reduces the dataset size to $|\mathcal{V}|l$, scales linearly with the number of landmarks $l \ll n$, and accelerates training on large graphs. It approximates all-pairs distribution by capturing the network's structural diversity. Selecting optimal landmarks is \textbf{NP-Hard}~\cite{potamias2009fast}. Random or heuristic selection provides an effective alternative.
This strategy simulates workloads with hotspots, and is the most widely adopted sampling approach in prior work~\cite{qi2020learning,chen2022ndist2vec}.

\noindent\textbf{Heuristic-based:} It designs custom sampling schemes~\cite{meng2024rgcndist2vec,zhao2022rne} using partitioning, clustering, or adaptive resampling of node pairs to balance coverage, by approximating all-pairs distribution, and efficiency, by selecting high-quality samples for training.

A unique aspect of shortest distance estimation is that the entire graph is known a priori.
This makes the dataset space \emph{fixed} and \emph{fully enumerable}, yet practically infeasible to exhaust due to the $\mathcal{O}(n^2)$ possible pairs, making the learning task non-trivial.
Synthetic datasets aim for representative coverage, whereas workload-driven datasets focus on realistic query patterns.
Our study covers both.

\subsection{Training and Optimization}
The optimization setup for learning shortest-path distance estimators includes the choice of loss functions and the overall training strategy, e.g., mini-batch learning in NN and GNNs.

\noindent\textbf{Loss Functions.}
Shortest-path distance estimation is formulated as a regression problem, where the goal is to minimize the discrepancy between predicted and true distances, $\hat{d}_i$ and $d_i$, over $N$ training samples. The objective is the Mean Squared Error (MSE), or $\ell_2$ loss,
$\mathcal{L}_{\text{MSE}} = \frac{1}{N} \sum_{i=1}^N (\hat{d}_i - d_i)^2,$
that penalizes larger deviations more strongly and works well when the training data is relatively clean. The Mean Absolute Error (MAE), or $\ell_1$ loss,
$\mathcal{L}_{\text{MAE}} = \frac{1}{N} \sum_{i=1}^N |\hat{d}_i - d_i|,$
is less sensitive to outliers but provides weaker gradients near zero. To combine the stability of $\ell_2$ with the robustness of $\ell_1$, several models (e.g., Vdist2vec-L) employ the Huber loss:
\[
\mathcal{L}_{\text{Huber}}(\hat{d}_i, d_i; \delta) =
\begin{cases}
\frac{1}{2} (\hat{d}_i - d_i)^2, & \text{if } |\hat{d}_i - d_i| \leq \delta,\\[4pt]
\delta (|\hat{d}_i - d_i| - \tfrac{1}{2}\delta), & \text{otherwise,}
\end{cases}
\]
where $\delta$ is a tunable threshold controlling the transition between quadratic and linear regimes.
The Smooth~$\ell_1$ variant~\cite{meng2024rgcndist2vec} is a special case of the Huber loss with $\delta{=}1$.

\noindent\textbf{Training.}
Neural models are optimized end-to-end using stochastic gradient-based optimizers, e.g., Adam~\cite{adam2014method} or SGD~\cite{kiefer1952stochastic,robbins1951stochastic}.
We adopt mini-batch training instead of full-batch optimization to balance convergence speed and memory efficiency on CPU and GPU. Each mini-batch contains sampled node pairs $(u,v)$, and only their embeddings are computed and updated.
This ensures scalability to large graphs by avoiding gradient computation over all node embeddings.
For models that learn node embeddings end-to-end, we use sparse embeddings coupled with sparse gradient updates (SparseAdam), so the optimizer step only touches the sampled nodes' embeddings, reducing per-epoch training time substantially.

For GNN models, training efficiency is improved by $L$-hop subgraph extraction. For each mini-batch of node pairs, we extract the induced subgraph formed by the union of their $L$-hop neighborhoods, where $L$ is the number of GNN layers. Message passing is performed only on this subgraph that possesses complete neighborhood information required for the mini-batch, while reducing unnecessary computation.
This procedure computes exact losses and gradients---it merely restricts message passing to relevant subgraphs---thereby preserving the correctness of training while substantially reducing computation and memory overhead. This makes GNN training time comparable to standard NN models.

\section{Experiments}
\label{sec:experiments}
To systematically evaluate ML-based distance indexes, we design a unified experimental framework that spans diverse datasets, models, and evaluation metrics. We examine performance along four axes: accuracy, query latency, preprocessing time, and model size.

Only a few original implementations are publicly available, and many are outdated, incomplete, or are built on different frameworks. Thus, we have reimplemented all models in PyTorch~\cite{paszke2019pytorch} and PyTorch Geometric~\cite{fey2019fast} using Python~3.12, adhering to their original architectures where available. This unified implementation ensures fair comparison across models by eliminating discrepancies arising from framework or dependency differences. All implementations are open-source to facilitate reproducibility and future research.
Experiments are run on a single-node SLURM-managed HPC cluster, where each job is allocated 16 CPU cores (AMD EPYC 7763), 32\,GB RAM, and one NVIDIA A100-SXM4-80GB GPU running CUDA~12.6 on Rocky Linux 9.7.
Inference latency/throughput benchmarks use ONNX Runtime (except Catboost) with TensorRT and CPU as execution providers for GPU and CPU, respectively. We fix the random seed across runs to ensure reproducibility and benchmarking.

\subsection{Datasets}
We use 13 real-world road-network datasets of varying scales and geographies, summarized in Table~\ref{tab:datasets}. Each dataset is modeled as an undirected, weighted graph, where vertices correspond to road intersections and edges represent road segments with travel distance as the weight. Geographic coordinates are available for all nodes, enabling methods that incorporate spatial features.
We construct workload-driven shortest distance queries from real-world trajectory datasets. We extract start--end locations from GeoLife~\cite{geolife-ref} trajectories for Beijing, taxi trip records for New York~\cite{nyc-taxi-ref} and Chicago~\cite{chicago-taxi-ref}, traffic camera trajectories for Jinan and Shenzhen~\cite{jinan-shenzhen-ref}, and taxi trajectories for Chengdu~\cite{chengdu-ref,wang2018will} and Shanghai~\cite{shanghai-ref}. For each trajectory, the origin and destination coordinates are map-matched to their nearest road-network vertices, and the resulting vertex pairs are used as shortest distance queries.
For the remaining datasets, we generate synthetic queries using all-pairs (Surat) and landmark-based sampling (FLA, E, W, CTR, USA) as described in Section~\ref{sec:query-dataset-prep}.

\begin{table}[t]
\caption{Real-world road network datasets. $k_{\text{max}}$, $k_{\text{avg}}$ are max., avg. node degrees and diam. is graph diameter (in km).
\label{tab:datasets}
}
\small
\setlength{\tabcolsep}{3pt}
\renewcommand{\arraystretch}{0.92}
\begin{tabular}{
    l  
    l  
    r  
    r  
    r  
    r  
    r  
    r  
}
\toprule
\textbf{Data}
 & \textbf{Region}
 & \textbf{Nodes}
 & \textbf{Edges}
 & \textbf{$k_{\text{max}}$}
 & \textbf{$k_{\text{avg}}$}
 & \textbf{diam.}
 & \textbf{Queries} \\
\midrule
\textbf{SU}~\cite{karduni2016protocol}
  & Surat        & 2.5k      & 3.6k      & 6     & 2.86  & 51   & 6.3M \\
\midrule
\textbf{JN}~\cite{jinan-shenzhen-ref}
  & Jinan        & 8.9k      & 14.1k     & 10    & 3.16  & 67   & 500k \\
\textbf{SZ}~\cite{jinan-shenzhen-ref}
  & Shenzhen     & 11.9k     & 18.9k     & 12    & 3.17  & 46   & 500k \\
\textbf{CD}~\cite{wang2018will}
  & Chengdu      & 17.6k     & 25.3k     & 11    & 2.88  & 223  & 500k \\
\textbf{BJ}~\cite{zheng2011geolife,geolife-ref}
  & Beijing      & 74.4k     & 103.4k    & 29    & 2.78  & 294  & 500k \\
\textbf{SH}~\cite{shanghai-ref}
  & Shanghai     & 74.9k     & 103.0k    & 9     & 2.75  & 127  & 500k \\
\textbf{NY}~\cite{nyc-taxi-ref}
  & NewYork      & 334.9k    & 445.9k    & 11    & 2.66  & 236  & 500k \\
\textbf{CH}~\cite{chicago-taxi-ref}
  & Chicago      & 386.5k    & 549.6k    & 9     & 2.84  & 191  & 500k \\
\midrule
\textbf{FLA}~\cite{9th-dimacs-ref}
  & Florida      & 1.1M     & 1.3M     & 8     & 2.51  & 1199 & 32.1M \\
\textbf{E}~\cite{9th-dimacs-ref}
  & Eastern US  & 3.6M     & 4.4M     & 9     & 2.42  & 2057 & 36.0M \\
\textbf{W}~\cite{9th-dimacs-ref}
  & Western US  & 6.3M     & 7.6M     & 9     & 2.41  & 3611 & 31.3M \\
\textbf{CTR}~\cite{9th-dimacs-ref}
  & Central US  & 14.1M    & 16.9M    & 8     & 2.41  & 3591 & 28.2M \\
\textbf{USA}~\cite{9th-dimacs-ref}
  & Full US     & 23.9M    & 28.9M    & 9     & 2.41  & 5586 & 23.9M \\
\bottomrule
\end{tabular}
\end{table}

\noindent\textbf{Data Preprocessing.}
For each network, we retain only the largest connected component to guarantee reachability among all queried vertex pairs. Node identifiers are re-indexed sequentially from 1 to $n$ for compactness. To enrich the workload, we augment unique real queries by perturbing endpoints with spatially proximate (8-hop neighborhood) nodes, resulting in 500K distance queries per workload-driven dataset.
The all-pairs (Surat) dataset uses all 6.3M node-pair queries, while the landmark-based DIMACS datasets (FLA, E, W, CTR, USA) use landmark-vertex pairs targeting $\sim$30M queries per dataset.
Data is split into 80\% training, 10\% validation, and 10\% testing.
Ground-truth distances come from non-ML distance index HC2L~\cite{HC2L-ref} that gives exact shortest-path distances.

\subsection{Models}
Evaluated models are grouped into five categories: (1) non-ML baselines, (2) NNs, (3) GNNs, (4) functional methods, and (5) gradient-boosted decision trees. Models' encoder-decoder configurations are in Table~\ref{tab:encoder-decoder-summary}. All learned models (except GeoDNN) use embedding dimension 64. Training budget is fixed per dataset group to match its query volume: 15 min for all-pairs (Surat), 5 min for workload-driven, and 60 min for landmark-based (DIMACS) datasets.
Full hyperparameter and training details are in Appendix~\ref{app:hyperparams}.

\vspace{3pt}
\noindent\textbf{\mcolor{mgBaseline}{Baselines (Non-ML).}}
We test against three baselines.
\textbf{Manhattan} estimates distance by $\ell_1$-norm over coordinates.
\textbf{Landmark$_{rn}$} and \textbf{Landmark$_{km}$} are landmark-based estimators using 64 landmarks from training vertices, selected randomly and nearest to $k$-Means centroids in Euclidean space, respectively.

\vspace{3pt}
\noindent\textbf{\mcolor{mgNN}{NNs.}}
We examine six neural architectures implemented following their respective papers (decoder architectures in Table~\ref{tab:encoder-decoder-summary}).
\textbf{GeoDNN}~\cite{jindal2017unified} uses geo-coordinates as input features.
\textbf{DistanceNN}~\cite{rizi2018shortest} and \textbf{EmbedNN}~\cite{qu2023learning} use pretrained Node2Vec embeddings with subtraction and averaging as fusion operators, respectively.
\textbf{Vdist2vec}~\cite{qi2020learning} and \textbf{Ndist2vec}~\cite{chen2022ndist2vec} learn node embeddings end-to-end from scratch.
Ndist2vec further uses a four-branch ensemble decoder.
\textbf{LandmarkNN} is an NN adaptation of CatBoost~\cite{jiang2021shortest} that replaces trees with fully-connected layers.

\vspace{3pt}
\noindent\textbf{\mcolor{mgGNN}{GNNs.}}
We evaluate three message-passing architectures using node coordinates as features.
\textbf{RGCNdist2vec}~\cite{meng2024rgcndist2vec} is the original GCN model with an $\ell_1$-norm decoder.
\textbf{RSAGEdist2vec} and \textbf{RGATdist2vec} are our variants that replace GCN layers with GraphSAGE~\cite{hamilton2017inductive} and graph attention (GAT)~\cite{velickovic2017graph}, respectively.

\vspace{3pt}
\noindent\textbf{\mcolor{mgFunctional}{Functional Methods.}}
We evaluate three embedding-based methods with functional decoders.
\textbf{Path2vec}~\cite{kutuzov2019making} learns randomly initialized embeddings using the modified loss in Equation~\eqref{eq:path2vec-loss}.
\textbf{ANEDA}~\cite{pacini2023aneda} learns embeddings initialized with pretrained Node2Vec.
\textbf{RNE}~\cite{zhao2022rne} trains hierarchical node embeddings.

\vspace{3pt}
\noindent\textbf{\mcolor{mgTree}{Gradient-Boosted Decision Trees (GBDT).}}
We evaluate \textbf{CatBoost}~\cite{jiang2021shortest} on a 64-dim. feature vector: 61 landmark distances, geo-coordinates, and a similarity feature. It is trained on CPU, where it converges faster and yields lower query latency than on GPU.
Unlike the original two-stage design, we use a single-stage decoder.

\subsection{Evaluation Metrics}
We use four primary metrics: approximation error, preprocessing time, query latency, and storage.
The first two are \emph{learning-dependent} quality metrics that improve with better optimization, whereas the latter two are \emph{architecture-dependent} efficiency metrics that remain fixed for a given model design.
Longer training may improve accuracy but cannot improve latency or storage—these are fixed by the architecture, not the optimization.
An exception is CatBoost whose size can vary slightly with additional training.
Measurement protocols for each metric are summarized below.

\noindent\textbf{Approximation Error (MRE).}
MRE quantifies how closely the model's predictions match the true shortest-path distances. We measure approximation quality using Mean Relative Error (MRE):
$\mathrm{MRE} = \frac{1}{N} \sum_{i=1}^{N} \frac{|\hat{d}_i - d_i|}{d_i}$.
MRE, also known as the Mean Absolute Percentage Error (MAPE), is scale-invariant and thus enables meaningful comparison across graphs with differing edge-weight or distance distributions. Thus, we adopt MRE instead of MAE.

\noindent\textbf{Precomputation Time (PT).}
PT measures offline cost (training or index construction).
We record the wall-clock time for a model to reach convergence or to exhaust a fixed training budget.
PT per epoch depends on the number of query samples $N$ and the model's architectural complexity $M$
(e.g., $M_{\text{GNN}} > M_{\text{MLP}}$ due to message-passing overhead in GNNs).
It does not directly depend on the underlying graph size, although $N$ may vary with graph scale depending on how the query workload is generated.
$\mathrm{PT}$ per epoch scales as $\mathcal{O}(MN)$, where $M$ reflects per-sample computation cost.

\noindent\textbf{Query Time (QT).}
QT is the query's average inference time (latency), measured as total inference time divided by the number of evaluated node pairs.
It depends on decoder complexity $M$ and embedding dim. $d$, roughly scaling as $\mathcal{O}(Md)$. For a fixed architecture, QT is \emph{independent} of training, graph size, or source-dest distance.

\noindent\textbf{Storage Overhead (MS).}
MS measures model size as the file size on disk post-training. It scales as $\mathcal{O}(dn)$ and remains constant for a fixed architecture, \emph{invariant} to training duration or quality.

\subsection{Results}

\begin{table*}[!h]
\caption{Comparison of test MRE (\%) among all methods and baselines across datasets. Lower values indicate better performance. \textbf{First}, \underline{second}, and \uwave{third} lowest values are highlighted.}
\label{tab:test-mre}
\small
\setlength{\tabcolsep}{4pt}
\renewcommand{\arraystretch}{0.92}
\begin{tabular}{
    l|                            
    r|                            
    rrrrrrr|                      
    *{5}{r}|                      
    r                             
}
\toprule
\multirow{2}{*}{\textbf{Model}}
 & \textbf{All-pairs}
 & \multicolumn{7}{c|}{\textbf{Workload-driven}}
 & \multicolumn{5}{c|}{\textbf{Landmark-based}}
 & \\
 & \textbf{SU}
 & \textbf{JN} & \textbf{SZ} & \textbf{CD} & \textbf{BJ} & \textbf{SH} & \textbf{NY} & \textbf{CH}
 & \textbf{FLA} & \textbf{E} & \textbf{W} & \textbf{CTR} & \textbf{USA}
 & \textbf{Mean $\pm$ Std} \\
\midrule
\multicolumn{15}{c}{\textbf{Baselines}} \\
\mcolor{mgBaseline}{Manhattan} & 13.49 & 9.60 & 11.46 & 11.43 & 9.13 & 9.88 & 20.39 & 9.67 & 14.16 & 15.00 & 10.98 & 14.88 & 15.43 & 12.73 $\pm$ 3.10 \\
\mcolor{mgBaseline}{Landmark$_{rn}$} & 10.43 & 46.37 & 31.70 & 29.68 & 54.04 & 37.82 & 17.43 & 9.03 & 7.33 & 7.08 & 6.71 & 6.95 & 12.00 & 21.28 $\pm$ 16.02 \\
\mcolor{mgBaseline}{Landmark$_{km}$} & 8.07 & 36.09 & 21.49 & 30.04 & 58.78 & 31.13 & 14.27 & 5.55 & 6.17 & 6.72 & 5.47 & 6.04 & 8.18 & 18.31 $\pm$ 15.81 \\
\midrule
\multicolumn{15}{c}{\textbf{Neural Network Methods}} \\
\mcolor{mgNN}{GeoDNN} & 5.47 & 6.65 & 6.24 & 7.14 & 7.19 & 5.28 & 4.33 & 2.02 & 1.75 & 1.14 & 0.89 & 0.51 & 0.45 & 3.77 $\pm$ 2.59 \\
\mcolor{mgNN}{DistanceNN} & 38.72 & 29.22 & 33.00 & 50.62 & 92.72 & 103.00 & 139.52 & 74.56 & 119.37 & 91.33 & 88.09 & 59.65 & 75.21 & 76.54 $\pm$ 32.40 \\
\mcolor{mgNN}{EmbedNN} & 9.43 & 15.29 & 15.61 & 22.11 & 40.21 & 53.98 & 71.86 & 22.83 & 125.90 & 99.76 & 96.80 & 65.50 & 86.07 & 55.80 $\pm$ 36.87 \\
\mcolor{mgNN}{Vdist2vec} & 2.11 & 8.26 & 6.48 & 10.96 & 37.35 & 43.94 & 31.37 & 5.94 & 2.95 & 5.37 & 17.31 & 66.07 & 86.64 & 24.98 $\pm$ 25.83 \\
\mcolor{mgNN}{Ndist2vec} & 2.19 & 7.09 & 5.79 & 10.34 & 34.10 & 40.59 & 32.57 & 32.60 & 5.63 & 5.51 & 30.01 & 61.43 & 82.22 & 26.93 $\pm$ 23.53 \\
\mcolor{mgNN}{LandmarkNN} & \textbf{0.89} & \textbf{3.10} & \textbf{2.63} & \textbf{2.34} & \underline{4.25} & \textbf{2.21} & \textbf{1.53} & \textbf{0.74} & \textbf{0.22} & \textbf{0.18} & \textbf{0.14} & \textbf{0.06} & \textbf{0.08} & \textbf{1.41 $\pm$ 1.32} \\
\midrule
\multicolumn{15}{c}{\textbf{Graph Neural Network Methods}} \\
\mcolor{mgGNN}{RGCNdist2vec} & 14.07 & 15.64 & 13.81 & 17.47 & 19.19 & 15.67 & 16.35 & 6.69 & 12.03 & 9.23 & 8.08 & 4.39 & 5.25 & 12.14 $\pm$ 4.72 \\
\mcolor{mgGNN}{RSAGEdist2vec} & 6.43 & 6.80 & 6.10 & 6.18 & 5.44 & 4.72 & 4.58 & 1.92 & 2.30 & 1.83 & 1.18 & 0.75 & 0.68 & 3.76 $\pm$ 2.26 \\
\mcolor{mgGNN}{RGATdist2vec} & 5.20 & 6.75 & 5.87 & 6.08 & \uwave{5.04} & \uwave{4.62} & 4.42 & 1.70 & 1.91 & 1.39 & 0.97 & 0.64 & 0.55 & 3.47 $\pm$ 2.21 \\
\midrule
\multicolumn{15}{c}{\textbf{Functional Methods}} \\
\mcolor{mgFunctional}{Path2vec} & 2.16 & 7.66 & 9.98 & 22.25 & 91.60 & 121.56 & 158.45 & 22.36 & 11.36 & 22.48 & 45.27 & 73.12 & 43.86 & 48.62 $\pm$ 46.94 \\
\mcolor{mgFunctional}{ANEDA} & 2.05 & \uwave{5.82} & 6.50 & 12.16 & 44.78 & 69.52 & 79.84 & 11.77 & 8.68 & 18.81 & 44.48 & 95.44 & 130.61 & 40.80 $\pm$ 39.72 \\
\mcolor{mgFunctional}{RNE} & \uwave{1.47} & 5.83 & \uwave{4.60} & \uwave{5.70} & 8.74 & 5.89 & \uwave{4.23} & \uwave{1.55} & \uwave{0.62} & \underline{0.28} & \underline{0.23} & \uwave{0.21} & \underline{0.09} & \uwave{3.03 $\pm$ 2.80} \\
\midrule
\multicolumn{15}{c}{\textbf{Tree-based Gradient Boosting Methods}} \\
\mcolor{mgTree}{CatBoost} & \underline{1.39} & \underline{3.29} & \underline{2.98} & \underline{2.87} & \textbf{3.08} & \underline{2.95} & \underline{1.94} & \underline{0.85} & \underline{0.46} & \uwave{0.38} & \uwave{0.28} & \underline{0.16} & \uwave{0.12} & \underline{1.60 $\pm$ 1.24} \\
\bottomrule
\end{tabular}
\end{table*}

\subsubsection{\textbf{Approximation Error (MRE)}}
\begin{figure*}[!t]
\centering
\includegraphics[width=0.99\linewidth]{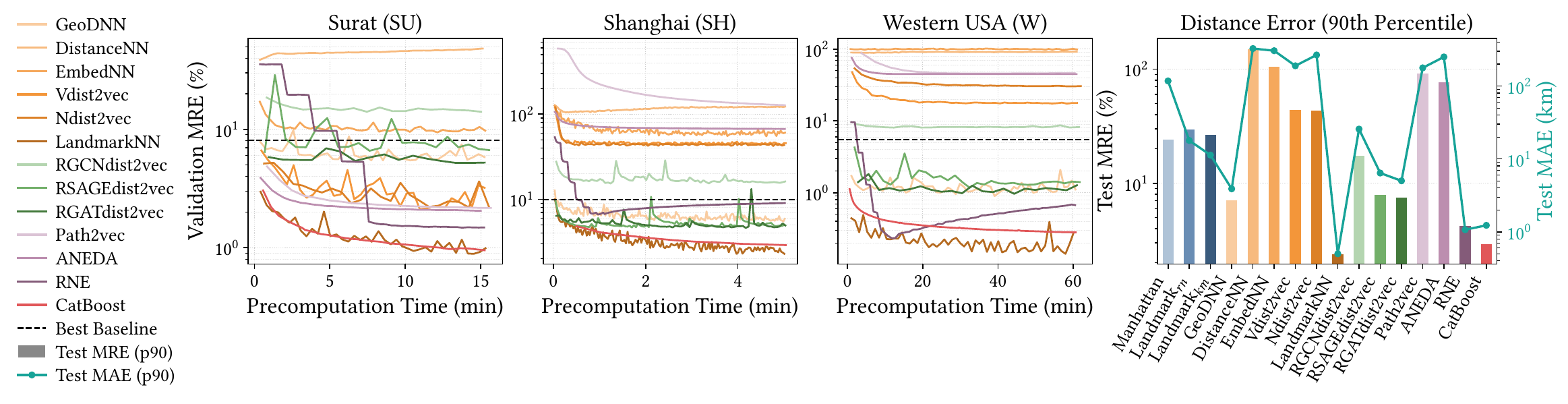}
\caption{Validation MRE (\%) as a function of precomputation time for three representative datasets (small, medium, and large road networks), and Test MRE (\%)/MAE (km) at the 90th percentile across all methods.
}
\label{fig:precomputation-time-vs-error}
\end{figure*}

Table~\ref{tab:test-mre} reports test MRE for the best model obtained using validation-set performance, under the applicable per-group training budget.
\textbf{LandmarkNN} achieves the lowest error on nearly all datasets (mean MRE 1.41\% $\pm$ 1.32\%), closely followed by \textbf{CatBoost} (1.60\% $\pm$ 1.24\%). We attribute this to their shared ability to exploit both global features (coordinates, landmark distances) and local features (similarity) while remaining robust across different datasets. LandmarkNN, our neural surrogate of CatBoost over the same feature space, benefits from a learnable decoder compared to CatBoost's tree-based approach.
\textbf{RNE} performs strongly among functional methods (3.03\% $\pm$ 2.80\%), especially on medium and large graphs.
Interestingly, within each dataset group, error tends to decrease as graphs grow larger for the best-performing method in each model category, suggesting that learned distance indexes scale favorably to larger networks.
The same three methods also top 90th-percentile error aggregated across datasets as shown in Figure~\ref{fig:precomputation-time-vs-error} (right).

Our workload-driven evaluation yields rankings that differ slightly from prior studies, which instead evaluate models on all (or uniformly sampled) node pairs~\cite{qi2020learning,chen2022ndist2vec,pacini2023aneda,meng2024rgcndist2vec} or landmark-generated~\cite{rizi2018shortest,zhao2022rne} synthetic queries---settings akin to our all-pairs (SU) results.
Real-world workloads do not weigh all vertices equally and are strongly biased towards short-range distances, where local features tend to dominate. In contrast, synthetic query sets tend to over-represent long-range distances that are well captured by global features, explaining the observed differences from prior results.

Among the baselines, \textbf{Manhattan} remains a strong reference on mid-sized workload-driven networks, but loses to landmark-based methods on all synthetic query datasets. To reflect workload awareness, we select landmarks from the training-query vertex set rather than from all vertices, which noticeably improves landmark accuracy. Centrality-based selections (betweenness, degree, closeness, etc.), advocated in earlier work for other graph families, offer no benefit on road networks. Simple random or $k$-means selections are competitive, but rarely surpass Manhattan.

Among NN methods, \textbf{LandmarkNN} leads consistently, due to its combination of global and local features. \textbf{Vdist2vec} and \textbf{Ndist2vec} perform well on smaller graphs but degrade on larger ones, where the simpler \textbf{GeoDNN} surpasses them while offering favorable memory and latency trade-offs (as discussed later). \textbf{EmbedNN} and \textbf{DistanceNN} underperform even the baselines across all datasets despite learning-rate tuning. We experiment with alternative fusion operators (sum, concatenation, etc.) and pretrained embeddings (e.g., LINE~\cite{tang2015line}), but the overall ranking remains unchanged.

Within message-passing GNNs, \textbf{RGATdist2vec} outperforms other variants, including the original GCN encoder~\cite{meng2024rgcndist2vec}. Adding edge weights provides no improvement. Thus, we report results on unweighted graphs. Attention mechanisms in GAT layers appear to implicitly capture effective weighting. For complex trajectories and topologies in Beijing (BJ) and Shanghai (SH), where most models exhibit higher error, GNNs remain comparatively more robust.

Among functional methods, \textbf{ANEDA} performs well on smaller graphs but deteriorates on larger ones. \textbf{RNE} is the most stable and often ranks behind LandmarkNN, suggesting that hierarchical structures integrate local and global effects in road networks.

\subsubsection{\textbf{Precomputation Time (PT)}}
All models are trained under their applicable per-group training budget.
Figure~\ref{fig:precomputation-time-vs-error} gives prediction error as a function of training time on 3 representative datasets (selected based on size), with other datasets showing similar trends. Models converge within their allocated budget, beyond which additional training yields diminishing returns.
\textbf{LandmarkNN}, the best-performing model, converges rapidly, followed by \textbf{CatBoost} and \textbf{RNE}. While the trend is less pronounced on smaller graphs, larger graphs more clearly distinguish competitive methods, as weaker ones plateau early or fail to improve. As expected, error generally decreases with additional training. However, RNE attains its best validation performance soon after the hierarchical training phase and begins to overfit thereafter.


\subsubsection{\textbf{Query Time (QT)}}
\label{sec:query-time}
Figure~\ref{fig:latency-storage} shows latency and throughput on CPU and GPU with batch size 1M, grouped by model category. For GPU, both metrics are reported with (end-to-end) and without (compute-only) data-transfer overhead. Setting a larger batch size significantly improves inference speed and overall throughput.
CPU values show mean $\pm$ std across all 13 datasets, each averaged over random 10M test queries.
GPU variance across datasets is negligible and not shown. Per-dataset and batch-size QT breakdowns are in Appendix~\ref{app:latency-ablation-study}.
Overall, GPU acceleration yields substantial gains of up to $\sim$200$\times$ for \textbf{LandmarkNN}, reducing latency to nanoseconds in learned indexes.
In general, latency increases with decoder complexity--from simple functional decoders to shallow and then deeper NNs.
ML models execute a fixed number of floating-point operations determined solely by network depth and dimensionality, yielding near-constant latency independent of graph size.

Within functional decoders, all methods show similar GPU latencies, governed mainly by embedding dimensionality (e.g., 2-dimensional coordinates for Manhattan vs. 64-dimensional inputs for landmark-based methods). GNN models follow the same trend since their node embeddings are precomputed and cached at inference time, so QT reflects only the decoder cost. At such low latencies (e.g., \textbf{RNE}), data-transfer overhead becomes the bottleneck while GPU computation remains extremely fast.

Among neural architectures, latency varies more widely depending on network depth, width, and activation functions. \textbf{LandmarkNN}, with its wider hidden layers, has the highest QT among NN methods, while simpler models, e.g., \textbf{GeoDNN} achieve faster inference.
Unlike other methods, \textbf{CatBoost}'s latency scales with tree count, causing high variance across datasets as tree counts grow under longer training budgets for larger graphs.
It does not benefit from GPU execution---neither during training nor inference---so only CPU results are reported. Extending GPU-parallelized tree inference remains a direction for future work.

\subsubsection{\textbf{Storage Overhead (MS)}}
\begin{figure*}[!t]
\centering
\includegraphics[width=0.9\linewidth]{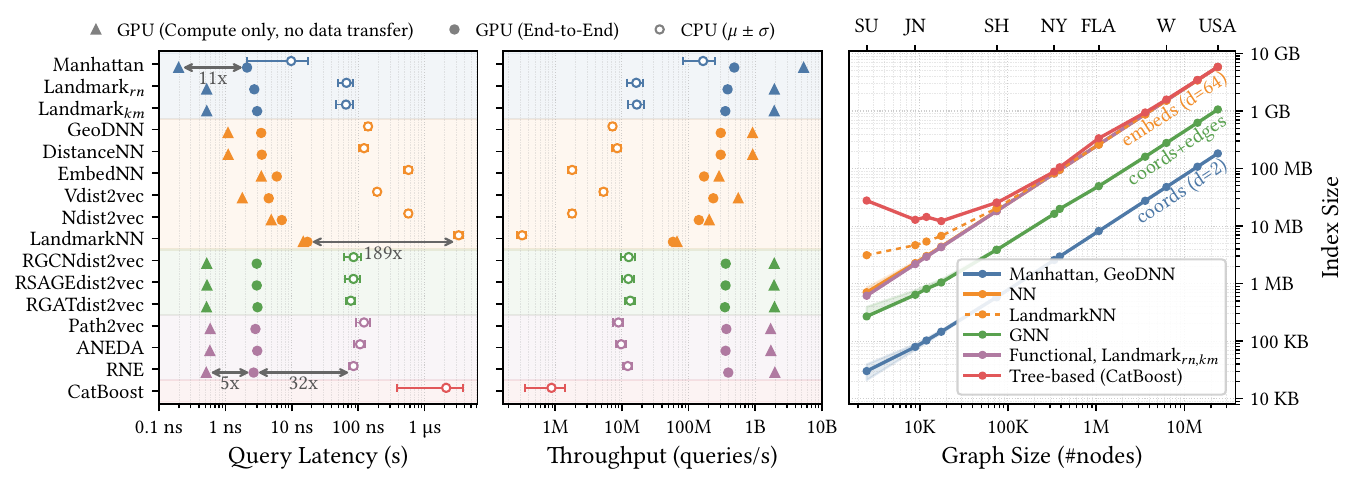}
\caption{Query efficiency (latency, throughput) on CPU vs. GPU and storage footprint (index size) across models.
}
\label{fig:latency-storage}
\end{figure*}

Figure~\ref{fig:latency-storage} plots index size versus graph size across all datasets and methods.
Index size is primarily determined by the encoder--specifically the embedding dimension and the number of stored embeddings (i.e., number of nodes).
Across datasets (left to right), index size grows roughly in proportion to the number of nodes, confirming the expected $\mathcal{O}(dn)$ scaling.

Across model categories, index size scales linearly with embedding dimensionality.
Methods storing only coordinates ($d{=}2$), e.g., \textbf{Manhattan} and \textbf{GeoDNN}, occupy minimal space.
GNN methods require slightly more storage due to additional edge lists for message passing.
Models with 64-dimensional embeddings (e.g., landmark baselines) consume about 32$\times$ more memory than coordinate-only models, consistent with the expected $\mathcal{O}(dn)$ scaling. \textbf{CatBoost} and \textbf{LandmarkNN} deviate slightly for smaller networks (${<}30$MB, due to their complex decoders) before converging to same linear trend.

\subsection{Discussion and Lessons Learned}
There is no single model that dominates in all settings; the optimal choice depends on the query workload and operational constraints.

\subsubsection*{\textbf{Workload Sensitivity}}
Our findings are particularly relevant for real-world query workloads, where short-range queries are more common than long-range ones. For long-range distance queries, even classical landmark-based baselines can achieve sub-1\% error. However, short-range predictions remain more challenging and benefit from more sophisticated learning-based distance indexes.

\subsubsection*{\textbf{Accuracy--Latency Trade-off}}
Under various workloads, \textbf{LandmarkNN} achieves the highest accuracy, closely followed by \textbf{CatBoost}, but incurs higher query latency. \textbf{RNE} instead offers the lowest latency with a modest increase in error.
\looseness=-1 With longer training budgets, LandmarkNN's accuracy can be further improved while maintaining low latency. Functional methods are preferred to NNs when latency is the bottleneck. GPU inference consistently yields up to two orders of magnitude speedup, making learned indexes feasible for real-time applications. Latency in ML-based indexes is deterministic, independent of graph size and source-destination distance.

\subsubsection*{\textbf{Storage Constraints}}
Storage for learned indexes scales predictably with $\mathcal{O}(dn)$.
Increasing training time improves accuracy but does not affect efficiency metrics (QT \& MS), except for tree-based models, e.g., CatBoost. Among learned methods, \textbf{LandmarkNN}, \textbf{RNE}, and \textbf{CatBoost} maintain comparable storage footprints as graph size grows. For memory-constrained environments, GeoDNN and GNN models are appealing due to their compact representations, despite slightly higher error rates.

\subsubsection*{\textbf{Generalization Capability}}
Methods that learn node embeddings directly (e.g., Vdist2vec, Ndist2vec) are transductive and cannot handle unseen nodes without retraining as each node must be observed during training for its embedding to be learned. In contrast, feature-driven or inductive approaches, e.g., GeoDNN, LandmarkNN, GNNs, and CatBoost, generalize naturally to unseen nodes and edges.

\subsubsection*{\textbf{Scalability Optimizations}}
Scaling to million-node road networks requires systems-level optimizations beyond model design. At inference, large batch sizes and a high-performance TensorRT runtime substantially improve throughput and latency. At training, sparse embeddings coupled with sparse gradient updates (SparseAdam) and GNN subgraph extraction substantially reduce per-epoch time without affecting correctness.

\subsubsection*{\textbf{Summary}}
In practice, \textbf{LandmarkNN} and \textbf{CatBoost} remain the most accurate choices, while \textbf{RNE} is the fastest.
Learned indexes consistently offer nanosecond-scale query latency and compact storage footprints while achieving acceptable approximation errors (typically within 3\%). However, learned embeddings introduce generalization and overfitting challenges that warrant further research.
Models trained on one road network may not transfer well to another. Inductive approaches such as GNNs generalize to unseen nodes and edges, making them a promising direction.

\section{Related Work}
\label{sec:related-work}

This paper relates to prior work on shortest-distance computation, distance indexing, learned distance approximation, and representation learning on graphs.

\noindent\textbf{Exact vs.\ Approximate Shortest Distance.}
Exact shortest distance methods, including classical graph algorithms~\cite{dijkstra1959note} and distance indexes~\cite{ch-ref,pathOracle-ref,hub-based-labeling2010-ref,H2H-ref}, guarantee optimal results but often incur substantial preprocessing cost or storage overhead on large road networks.
Specialized indexes have also been proposed for spatial queries such as k-nearest neighbor search~\cite{papadias2003query,kolahdouzan2004voronoi,hu2006distance}.
Approximate approaches relax exactness guarantees to improve efficiency and include both non-ML techniques~\cite{sketch-do2-ref,landmarkA*-ref,billionNodeDO-ref,DO-2009-ref,potamias2009fast,takes2014adaptive} and ML-based~\cite{jindal2017unified,rizi2018shortest,qu2023learning, qi2020learning,chen2022ndist2vec,kutuzov2019making,pacini2023aneda,zhao2022rne,jiang2021shortest,meng2024rgcndist2vec} methods that learn compact representations of the distance function.
These approaches expose trade-offs among accuracy, query latency, preprocessing time, and storage, that are central to this benchmark.

\noindent\textbf{Related Problems and Graph Domains.}
Learning on road networks has been studied in related problems, e.g., travel time estimation~\cite{huang2022context,ghosal2019travel,zhang2018deeptravel}, next-location and destination prediction~\cite{wu2020learning}, traffic flow forecasting~\cite{abadi2014traffic}, and road segment classification~\cite{wu2020learning,chen2021robust}.
These problems are closely related to representation learning on general purpose graphs, where a large body of work has proposed methods based on random walks (e.g., Node2vec~\cite{grover2016node2vec}, LINE~\cite{tang2015line}, DeepWalk~\cite{perozzi2014deepwalk}), as well as GNNs, e.g., GCN~\cite{kipf2016semi}, GraphSAGE~\cite{hamilton2017inductive}, GAT~\cite{velickovic2017graph}.
Moreover, distance or similarity learning has been explored on other graph domains, including social networks~\cite{gubichev2010fast,takes2014adaptive,rizi2018shortest,qu2023learning}, communication networks~\cite{suzuki2020estimating,jiang2022graph}, knowledge graphs~\cite{kutuzov2019making,cai2024survey}, and biological networks~\cite{muzio2021biological}.

\section{Conclusion and Future Work}
\label{sec:conclusion}
We present the first unified empirical evaluation of ML-based distance indexes for road networks, benchmarking ten representative models across 13 real-world datasets. Our study compares accuracy, query latency, preprocessing time, and storage, revealing clear trade-offs: LandmarkNN delivers the highest accuracy, closely followed by CatBoost, while functional methods, e.g., RNE, offer the best latency--accuracy balance. This benchmark establishes a foundation for advancing learned distance indexes. Promising directions for future work include developing architectures and training procedures for improved accuracy, building update-efficient models for dynamic road networks, strengthening GNN-based approaches, and curating richer workload-driven query benchmarks.

\begin{acks}
The authors declare the use of generative AI tools (ChatGPT, Claude Code) in the preparation of this manuscript. These tools assisted with refining prose, formatting LaTeX, and generating code. All outputs were reviewed, edited, and validated by the authors.
\end{acks}

\clearpage

\balance

\bibliographystyle{ACM-Reference-Format}
\bibliography{references}

\clearpage
\appendix

\section{Hyperparameter and Training Details}
\label{app:hyperparams}

\noindent\textbf{NNs.}
Models use ReLU activation in hidden layers. Each network ends with a single scalar output neuron, employing either a sigmoid (scaled by the maximum distance) or ReLU (without scaling) activation; DistanceNN instead uses a SoftPlus output and dropout ($p{=}0.4$). Models are trained with MSE loss to predict shortest distance. The embedding dimension is 64.
Training is conducted for 15 minutes on all-pairs (Surat), 5 minutes on workload-driven, and 60 minutes on landmark-based (DIMACS) datasets, using the Adam optimizer~\cite{adam2014method} (learning rate 0.01, except 0.0003 for LandmarkNN that requires a slower rate) and a 1024 batch size. Models with a learnable embedding table (e.g., Vdist2vec, Ndist2vec, ANEDA, Path2vec, RNE) use SparseAdam in place of dense Adam. SparseAdam updates only the sampled nodes' embeddings, improving training efficiency.
For Vdist2vec and Ndist2vec, the node-embedding tables and linear layer weights are initialized by a truncated normal distribution ($\mu{=}0$, $\sigma{=}0.01$), and biases are 0~\cite{qi2020learning,chen2022ndist2vec}.
For all other models, we retain PyTorch's defaults: Kaiming uniform~\cite{he2015delving} for linear layers and Normal $\mathcal{N}(0,1)$ for embedding tables.

We obtain Node2Vec embeddings by performing 10 unbiased random walks of Length 80 per node on the undirected graph ($p{=}q{=}1$) and training a skip-gram model with Window Size 10 for 1 epoch at Learning Rate 0.05 to produce 64-dimensional embeddings.

\noindent\textbf{GNNs.}
GNNs use Leaky-ReLU activations and are trained by the Smooth~$\ell_1$ loss under the same optimizer and training configuration as the NN models.

\noindent\textbf{Functional Methods.}
ANEDA and Path2vec use a learning rate of 0.03, while RNE uses 0.003 by default, reduced to 0.001 on five large road networks (NewYork, Chicago, FLA, E, W), and further reduced to 0.0003 on the two largest, CTR and USA.
Learning rates are determined by hyperparameter tuning, with all other training settings identical to the NN models.
RNE's embedding table is initialized as $\mathrm{Uniform}(-1.5, 1.5)$, deviating from the PyTorch default described above.

\noindent\textbf{CatBoost.}
CatBoost is trained using RMSE loss, Learning Rate 0.3, in chunks of 500 trees until the applicable per-group training budget is reached — actual tree counts trained range from roughly 5K (workload-driven), to 15.5K (Surat, all-pairs), to 61K (DIMACS).

\section{Query Time (QT) Ablation Study}
\label{app:latency-ablation-study}

\begin{figure*}[!b]
\centering
\includegraphics[width=0.9\linewidth]{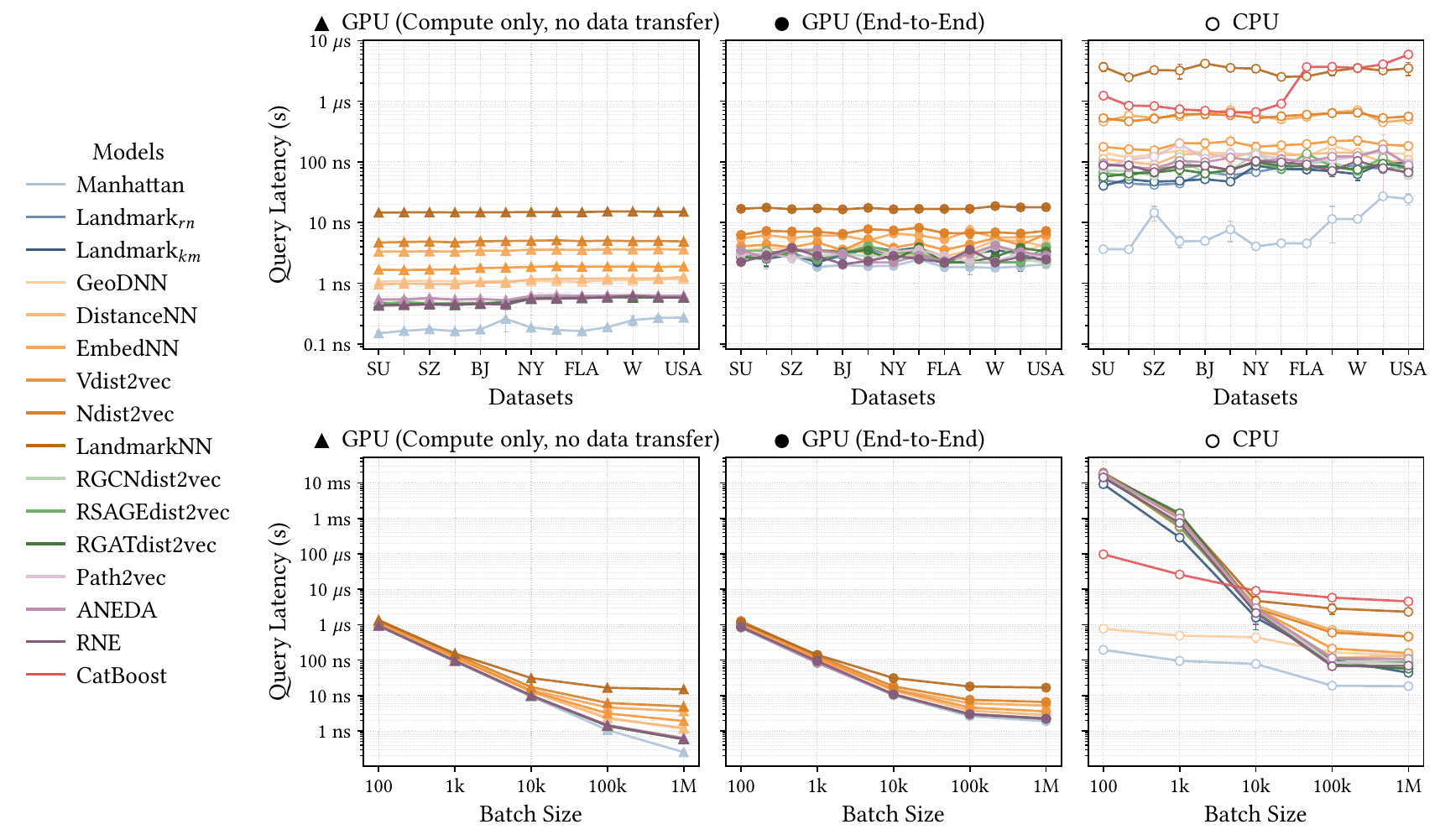}
\caption{Query latency vs. datasets (batch size 1M, top) and vs. batch size on USA (bottom), on GPU and CPU.}
\label{fig:latency-vs-dataset}
\end{figure*}

\noindent\textbf{Effect of Dataset Scale.}
Figure~\ref{fig:latency-vs-dataset} (top) plots query latency (batch size 1M) across all 13 datasets, ordered by graph size. GPU latency (compute-only and end-to-end) is flat across dataset scale, confirming it is governed by decoder complexity, not graph size. CPU latency is noisier and, for \textbf{CatBoost}, it grows with dataset size as tree count grows under longer training budgets.

\noindent\textbf{Effect of Batch Size.}
Figure~\ref{fig:latency-vs-dataset} (bottom) plots query latency vs. batch size on the largest dataset (USA). Latency falls monotonically as batch size grows, most sharply on CPU, confirming that larger batches amortize fixed per-call overhead and drive the throughput gains reported in Section~\ref{sec:query-time}.

\end{document}